\documentclass[graybox]{svmult}

\usepackage{xcolor}
\newcommand\hl[1]{%
  \bgroup
  \hskip0pt\color{red!80!black}%
  #1%
  \egroup
}

\usepackage{type1cm}        % activate if the above 3 fonts are
\usepackage{makeidx}         % allows index generation
\usepackage{graphicx}        % standard LaTeX graphics tool
\usepackage{multicol}        % used for the two-column index
\usepackage[bottom]{footmisc}% places footnotes at page bottom

\usepackage{newtxtext}       % 
\usepackage{newtxmath}       % selects Times Roman as basic font
\usepackage{makecell}

\makeindex             % used for the subject index
\begin{document}

\title*{Fourth-Order Anisotropic Diffusion for Inpainting and Image Compression}
\author{Ikram Jumakulyyev and Thomas Schultz}
\institute{Ikram Jumakulyyev \and Thomas Schultz \at University of Bonn, B-IT and Department of Computer Science II, Endenicher Allee 19A, 53115 Bonn, Germany \email{ijumakulyyev@cs.uni-bonn.de; schultz@cs.uni-bonn.de}}
\maketitle

\abstract{Edge-enhancing diffusion (EED) can reconstruct a close approximation of an original image from a small subset of its pixels. This makes it an attractive foundation for PDE based image compression. In this work, we generalize second-order EED to a fourth-order counterpart. It involves a fourth-order diffusion tensor that is constructed from the regularized image gradient in a similar way as in traditional second-order EED, permitting diffusion along edges, while applying a non-linear diffusivity function across them. We show that our fourth-order diffusion tensor formalism provides a unifying framework for all previous anisotropic fourth-order diffusion based methods, and that it provides additional flexibility. We achieve an efficient implementation using a fast semi-iterative scheme. Experimental results on natural and medical images suggest that our novel fourth-order method produces more accurate reconstructions compared to the existing second-order EED.}

\section{Introduction}
\label{sec:1}
The increased availability and resolution of imaging technology, including digital cameras and medical imaging devices, along with advances in storage capacity and transfer bandwidths, have led to a proliferation of large image data. This makes image compression an important area of research. \index{Image compression techniques} Image compression techniques can be divided into two main groups: Lossy and lossless compression. \index{ Lossless compression} Lossless compression techniques permit restoration of the full, unmodified image data, which however limits the achievable compression rates. Our work is concerned with \index{lossy compression} lossy compression, which achieves much higher compression rates by replacing the original image with an approximation that can be stored more efficiently.

We continue a line of research that has explored the use of \index{Partial Differential Equations (PDEs)} Partial Differential Equations (PDEs) for lossy image compression \cite{galic2005towards, galic2008image, schmaltz2009beating, mainberger2009edge}. This approach is based on storing only a small subset of all pixels, and interpolating between them in order to restore the remaining ones. There is a strong similarity between that interpolation process and \index{image inpainting} image inpainting, whose goal it is to reconstruct missing or corrupted parts of an image. \index{PDE-based methods} PDE-based methods for image inpainting and compression are inspired by the physical phenomenon of \index{heat transport} heat transport. It is described by the heat diffusion equation 
	\begin{equation} \label{eq:heatModel}
		{\partial_t u} = \mathrm{div}(D \cdot \nabla u) \;  ,
	\end{equation}
which relates temporal changes in a heat concentration $\partial_t u$ to the divergence of its spatial gradient $\nabla u$. When diffusion takes place in an isotropic medium, the diffusivity $D$ is a scalar that determines the rate of heat transfer. In an anisotropic medium, heat spreads out more rapidly in some directions than in others. In those cases, $D$ is a \index{diffusion tensor} diffusion tensor, i.e., a symmetric matrix that encodes this directional dependence.

When applied to image processing, the gray value at a certain location is interpreted as the heat concentration $u$. In diffusion-based \index{image inpainting} image inpainting, Equation~(\ref{eq:heatModel}) is used to propagate information from the known pixels, whose intensity is fixed, to the unknown pixels which will ultimately reach a steady state in which their intensity is determined by their surrounding known pixels. In this sense, Equation~(\ref{eq:heatModel}) has a filling-in effect that can be exploited for image compression.

Different choices of the \index{diffusivity function} diffusivity function $D$ lead to different kinds of diffusion. \index{Linear diffusion} Linear diffusion \cite{iijima1962basic} and \index{nonlinear diffusion} nonlinear diffusion \cite{perona1990scale} were widely used for image smoothing and image enhancement. \index{Edge structures} Edge structures in images can be enhanced by employing a diffusion tensor which allows diffusion in the direction perpendicular to the local gradient, while applying a nonlinear diffusivity function along the gradient direction. This idea has led to the development of \index{edge-enhancing diffusion (EED)} anisotropic nonlinear edge-enhancing diffusion (EED) \cite{weickert1998anisotropic}. Among the six variants that were evaluated for image compression by Gali{\'c} et al. \cite{galic2008image}, EED led to the most accurate reconstructions. Subsequently, this idea was applied to three-dimensional data compression \cite{peter2013three}, and combined with motion compensation in order to obtain a framework for video compression \cite{andris2016proof}. When combined with a suitable scheme for selecting and storing the preserved pixels, a few additional optimizations, and at sufficiently high compression rates, anisotropic diffusion has been shown to beat the quality even of JPEG2000 \cite{schmaltz2014understanding}.

In this paper, we introduce a novel \index{fourth-order PDE} fourth-order PDE that generalizes second-order EED, and achieves even more accurate reconstructions. 
We build on prior works that proposed fourth-order analogs of the diffusion equation, and used them for image processing \cite{scherzer1998denoising,you2000fourth,lysaker2003noise,didas2009properties,hajiaboli2011anisotropic,li2013two}. In particular, we extend a work by Gorgi Zadeh et al.\ \cite{zadeh2017multi}, who introduced the idea of steering \index{anisotropic fourth-order diffusion} anisotropic fourth-order diffusion with a \index{fourth-order diffusion tensor} fourth-order diffusion tensor. However, their method focuses on the \index{curvature enhancement} curvature enhancement property of nonlinear fourth-order diffusion \cite{didas2009properties} in order to better localize ridge and valley structures. Deriving a suitable PDE for image inpainting requires a different definition of the diffusion tensor, more similar to the one in edge-enhancing diffusion \cite{weickert1998anisotropic}. Two anisotropic fourth-order PDEs for inpainting were previously introduced by Li et al. \cite{li2013two}. However, they only apply them to image restoration tasks in which small parts of an image are missing (such as in Figure \ref{fig:imgRestThin}), not to the reconstruction from a small subset of pixels. Moreover, we demonstrate that the fourth-order diffusion tensor based framework is more general in the sense that it can be used to express anisotropic fourth-order diffusion as it was described by Hajiaboli \cite{hajiaboli2011anisotropic} or by Li et al. \cite{li2013two}, while providing additional flexibility.

\section{Background and Related Work}
\label{sec:2}

We will now formalize the above-mentioned idea of \index{diffusion-based inpainting} diffusion-based inpainting (Section~\ref{subsec:2}), and review two concepts that play a central role in our method: \index{Anisotropic nonlinear diffusion} Anisotropic nonlinear diffusion (Section~\ref{subsec:nonlinear}) and \index{fourth-order diffusion} fourth-order diffusion (Section~\ref{subsec:fourth-order}). Further details can be found in works by Gali{\'c} et al.\ \cite{galic2008image} and Weickert \cite{weickert1998anisotropic}, respectively. Finally, we provide additional context with a brief discussion of alternative approaches to \index{image compression} image compression (Section~\ref{subsec:alternative-compression}).

\subsection{Diffusion-based Inpainting}
\label{subsec:2}

In order to apply Equation~(\ref{eq:heatModel}) to \index{image smoothing} image smoothing, we have to restrict it to the image domain $\Omega$, and specify the behavior along its boundary $\partial\Omega$. It is common to assume that no heat is transferred through that boundary (homogeneous \index{Neumann boundary condition} Neumann boundary condition). Moreover, the positive real line $(0,\infty)$ is typically taken as the time domain. The resulting PDE can be written as
\begin{equation} \label{eq:smoothingModel}
\begin{split}
	{\partial_t u} &= \mathrm{div}(D \cdot \nabla u), \quad \Omega \times (0,\infty)\ , \\
	{\partial_n u} &= 0, \quad \partial\Omega \times (0,\infty) \ ,
%        u &= f, \quad \Omega \times 0
\end{split}
\end{equation}
where $n$ is the normal vector to the boundary $\partial\Omega$. The original image $f:\Omega\rightarrow\mathbb{R}$ is used to specify an initial condition $u=f$ at $t=0$. For increasing diffusion time $t$, $u$ will correspond to an increasingly smoothed version of the image.

In \index{image inpainting} image inpainting, we know the pixel values on a subset $K\subset \Omega$ of the image, and aim to reconstruct plausible values in the unknown regions. A diffusion-based model for inpainting can be derived from the one for smoothing, by modeling the set of locations at which the pixel values are known with \index{Dirichlet boundary condition} Dirichlet boundary conditions. In this case, $f:K\rightarrow\mathbb{R}$ will be used to model the known values. In inpainting-based image compression, $K$ will consist of a small fraction of the pixels in the original image. With this, we obtain the following model for inpainting:
\begin{equation} \label{eq:inpaintingModel}
\begin{split}
	{\partial_t u} &= \mathrm{div}(D \cdot \nabla u), \quad \Omega \backslash K \times (0,\infty)\ , \\
	{\partial_n u} &= 0, \quad \partial\Omega \times (0,\infty) \ , \\
	u &= f, \quad K\times [0,\infty)
\end{split}
\end{equation}

In this case, the diffusion process spreads out the information from the known pixels to their spatial neighborhood. For time $t\rightarrow\infty$, \index{image smoothing} image smoothing and inpainting both converge to a steady state, i.e., $\lim_{t\rightarrow\infty} \partial_t u=0$. However, the steady-state of smoothing is trivial ($u$ approaches a constant image with average gray value), while the Dirichlet boundary conditions in the inpainting case ensure a non-trivial steady-state, which is taken as the final inpainting result: $u_\text{inpainted} = \lim_{t\rightarrow\infty}u$

\subsection{From Linear to Anisotropic Nonlinear Diffusion}
\label{subsec:nonlinear}
So far, we assumed that the diffusion coefficient $D$ is a scalar and constant, independent from the location within the image. This results in an inpainting model based on \index{linear homogeneous diffusion} linear homogeneous diffusion \cite{iijima1962basic}. With $D = 1$, it can be written as
\begin{equation} \label{eq:linHomModel}
\begin{split}
	{\partial_t u} &= \Delta u, \quad \Omega \backslash K \times (0,\infty)\ .% \\
%	{\partial_n u} &= 0, \quad \partial\Omega \times (0,\infty) \ , \\
%	u &= f, \quad K\times [0,\infty) \ .
\end{split}
\end{equation}

In this and all remaining equations in this section, the same boundary conditions are assumed as specified in Equation~(\ref{eq:inpaintingModel}).
Despite its simplicity, it has been demonstrated that using this inpainting model for \index{image compression} image compression can already beat the JPEG standard when applied to cartoon-like images, and selecting the retained pixels to be close to \index{image edges} image edges \cite{mainberger2009edge}.

When the diffusion coefficient is a scalar but depends on $u$, i.e., $D = g(u)$, then we call the model inpainting based on nonlinear isotropic diffusion \cite{perona1990scale}. A common variant is to make $D$ depend on the local gradient magnitude, i.e.,
\begin{equation} \label{eq:nonIsoModel}
\begin{split}
	{\partial_t u} &= \mathrm{div}(g(||\nabla u_\sigma||^2) \nabla u), \quad \Omega \backslash K \times (0,\infty)\ , \\
%	{\partial_n u} &= 0, \quad \partial\Omega \times (0,\infty) \ , \\
%	u &= f, \quad K\times [0,\infty) \ ,
\end{split}
\end{equation}
where $g$ is a decreasing nonnegative diffusivity function, e.g., the Charbonnier diffusivity
\begin{equation} \label{eq:CharDiff}
	g(s^2) = \frac{1}{\sqrt{1+\frac{s^2}{\lambda^2}}} \  ,
\end{equation} 
and $\lambda$ is a \index{contrast parameter} contrast parameter separating low from high diffusion areas \cite{charbonnier1997deterministic}. In order to localize edges better and to make the problem well-posed, the image is pre-smoothed with a Gaussian before taking its gradient, i.e., $g(||\nabla u_{\sigma}||^2)$ is used instead of $g(||\nabla u||^2)$ \cite{catte1992image}.

In the above-discussed models, the diffusion occurs only in the gradient direction. This can be changed by replacing the scalar diffusivity with a second-order diffusion tensor, i.e., a symmetric positive definite matrix. This is the basis of \index{anisotropic nonlinear diffusion} anisotropic nonlinear diffusion \cite{weickert1998anisotropic},
\begin{equation} \label{eq:EEDModel}
	{\partial_t u} = \mathrm{div}(\mathbf{D} \cdot\nabla u), \quad \Omega \backslash K \times (0,\infty)\ .
\end{equation}  

In edge-enhancing diffusion (EED), the diffusion tensor $\mathbf{D}$ is defined as
\begin{equation} \label{eq:diffTensor_EED}
	\mathbf{D} = g(||\nabla u_{\sigma}||^2) \cdot \mathbf{v}_1 \mathbf{v}_1^\mathrm{T} + 1 \cdot \mathbf{v}_2 \mathbf{v}_2^\mathrm{T},
\end{equation} 
where $\mathbf{v}_1 = \frac{\nabla u_{\sigma}}{||\nabla u_{\sigma}||_2}$ and $\mathbf{v}_2 = \frac{\nabla u_{\sigma}^{\perp}}{||\nabla u_{\sigma}||_2}$. This means that diffusion across the edge ($\mathbf{v}_1$) is decreased depending on the gradient magnitude, while diffusion along the edge ($\mathbf{v}_2$) is allowed. Examples of EED based inpainting are included in our experimental results. In general, EED based inpainting results in better interpolated images than linear homogeneous or nonlinear isotropic PDEs. This makes it a current state-of-the-art choice for \index{PDE-based image compression} PDE-based image compression.

\subsection{From Second to Fourth Order Diffusion}
\label{subsec:fourth-order}
All models discussed above, as well as several others that have been proposed for inpainting \cite{zhang2015partial}, share a common property: They rely on second order PDEs. In image denoising, higher-order PDEs have a long history, going back to work by Scherzer \cite{scherzer1998denoising}. You and Kaveh \cite{you2000fourth} propose fourth-order PDEs as a solution to the so-called staircasing problem that arises in edge-enhancing second-order PDEs, such as the filter proposed by Perona and Malik \cite{perona1990scale}: While the second-order \index{Perona-Malik equation} Perona-Malik equation creates visually unpleasant step edges from continuous variations of intensity, corresponding fourth-order methods move these discontinuities into the gradients, where they are less noticeable to the human eye \cite{greer2003h}. Subsequently, other fourth-order PDE-based models have been introduced, and have mostly been applied for denoising \cite{lysaker2003noise, hajiaboli2009self, hajiaboli2011anisotropic}.

For a specific family of \index{higher-order diffusion filters} higher-order diffusion filters, Didas et al.\ \cite{didas2009properties} have shown that, in addition to preserving average gray value, they also preserve higher moments of the initial image. Moreover, depending on the diffusivity function, they can lead to adaptive forward and backward diffusion, and therefore to the enhancement of image features such as curvature. Gorgi Zadeh et al. \cite{zadeh2017multi} made use of this property in order to enhance ridges and valleys, by steering fourth-order diffusion with a fourth-order diffusion tensor. Our work adapts their method in order to achieve accurate inpainting and reconstruction from a small subset of pixels.

\subsection{Alternative Approaches to Image Compression}
\label{subsec:alternative-compression}

The dominant \index{lossy image compression} lossy image compression techniques today are JPEG and JPEG2000. They are based on the discrete cosine transform (DCT) and wavelet transform, respectively. However, they are not sensitive to the geometry of an image, i.e. those standards are not tailored to their geometrical behavior \cite{IEEE_TIP-Cagnazzo2007}. Especially, the JPEG standard involves dividing the image into small square blocks. This can cause a degradation called \index{blocking effect} ``blocking effect'' \cite{thayammal2013review}, and can result in unsatisfactory reconstructions especially at high compression rates.

It is an ongoing research trend to apply machine learning methods to image compression, such as convolutional and recurrent neural networks \cite{TheisSCH17, BalleLS17, TodericiOHVMBCS15}. Learning based approaches tend to perform very well on the specific class of images on which they were trained, but require a huge amount of data. For example, Toderici et al. \cite{toderici2017full} used for training a dataset of 6 million $1280\times 720$ images taken from the web.

%-------------------------------------------------------------------------

\section{Method}
\label{sec:3}

We will now introduce our novel PDE (Section~\ref{subsec:our-pde}), investigate its relationship to previously proposed \index{anisotropic fourth-order diffusion} anisotropic fourth-order diffusion (Section~\ref{subsec:comparison}), and comment on our chosen \index{discretization} discretization, as well as \index{numerical stability} numerical stability (Section~\ref{subsec:implementation}).

\subsection{Anisotropic Edge-Enhancing Fourth Order PDE}
\label{subsec:our-pde}

Our fourth-order PDE builds on a model that was proposed by Gorgi Zadeh et al.\ \cite{zadeh2017multi} for ridge and valley enhancement. It can be stated concisely using Einstein notation, where summation is implied for indices appearing twice in the same expression:
\begin{equation}
\label{eq:ts-fourth-order-diffusion}
\partial_{t}u = -\partial_{ij}\left[\mathcal{D}(\mathbf{H}_{\rho}(u_{\sigma})):\mathbf{H}(u)\right]_{ij}
\end{equation}
In this equation, $\mathbf{H}(u)$ denotes the \index{Hessian matrix} Hessian matrix of image $u$. The ``double dot product'' $\mathbf{T}=\mathcal{D}:\mathbf{H}$ indicates that matrix $\mathbf{T}$ is obtained by applying a linear map $\mathcal{D}$ to $\mathbf{H}$, and the square bracket notation $[\mathbf{T}]_{ij}$ indicates taking the $(i,j)$th component:
\begin{equation}
  \label{eq:double-dot}
  [\mathbf{T}]_{ij} = \left[\mathcal{D}(\mathbf{H}_{\rho}(u_{\sigma})):\mathbf{H}(u)\right]_{ij} = \left[\mathcal{D}(\mathbf{H}_{\rho}(u_{\sigma}))\right]_{ijkl} \left[\mathbf{H}(u)\right]_{kl}
\end{equation}
Since $\mathcal{D}$ maps matrices to matrices, it is a \index{fourth-order tensor} fourth-order tensor. Since its role is analogous to the second-order diffusion tensor in Equation~(\ref{eq:EEDModel}), it is referred to as a fourth-order diffusion tensor.

The \index{diffusion tensor} diffusion tensor $\mathcal{D}$ in Equation~(\ref{eq:ts-fourth-order-diffusion}) is a function of the local normalized Hessian $\mathbf{H}_{\rho}(u_{\sigma})$ which contains the information that is relevant to achieve \index{curvature enhancement} curvature enhancement. For image inpainting, we propose to instead steer the fourth-order diffusion in analogy to edge-enhancing diffusion, i.e., as a function of the \index{structure tensor} structure tensor $\mathbf{J}(u_\sigma)$, which is obtained from image $u$ after Gaussian pre-smoothing with bandwidth $\sigma$. We construct our fourth order diffusion tensor $\mathcal{D}$ from its eigenvalues $\mu_i$ and eigentensors $\mathbf{E}_i$ via the spectral decomposition:
\begin{equation} \label{eq:faod_diffTensor}
	\mathcal{D}\left(\mathbf{J}(u_\sigma)\right) = \mu_1 \mathbf{E}_1 \otimes \mathbf{E}_1 + \mu_2 \mathbf{E}_2 \otimes \mathbf{E}_2 + \mu_3 \mathbf{E}_3 \otimes \mathbf{E}_3 + \mu_4 \mathbf{E}_4 \otimes \mathbf{E}_4
\end{equation}
The eigenvalues and eigentensors are defined as
\begin{equation} \label{eq:foad_diffTensor_entries}
\begin{split}
	\mu_1 &= g(\lambda_1), \ \mathbf{E}_1=\mathbf{v}_1 \otimes \mathbf{v}_1 \ , \\
	\mu_2 &= 1, \ \mathbf{E}_2=\mathbf{v}_2 \otimes \mathbf{v}_2 \ , \\
	\mu_3 &= \sqrt{g(\lambda_1)}, \ \mathbf{E}_3=\frac{1}{\sqrt{2}}(\mathbf{v}_1 \otimes \mathbf{v}_2 + \mathbf{v}_2 \otimes \mathbf{v}_1) \ , \\
	\mu_4 &= 0, \ \mathbf{E}_4=\frac{1}{\sqrt{2}}(\mathbf{v}_1 \otimes \mathbf{v}_2 - \mathbf{v}_2 \otimes \mathbf{v}_1) \ ,
\end{split}
\end{equation}
where $g$ is a nonnegative decreasing diffusivity function, $\lambda_i$ and $\mathbf{v}_i$ are eigenvalues and eigenvectors of the structure tensor $\mathbf{J}(u_\sigma)=\nabla u_{\sigma}\nabla u_{\sigma}^\mathrm{T}$, i.e., $\lambda_1 = ||\nabla u_{\sigma}||_2^2$, $\mathbf{v}_1 = \frac{\nabla u_{\sigma}}{||\nabla u_{\sigma}||_2}$ and $\lambda_2 = 0$, $\mathbf{v}_2 = \frac{\nabla u_{\sigma}^{\perp}}{||\nabla u_{\sigma}||_2}$. The above-defined eigentensors are orthonormal with respect to the dot product $\mathbf{A}:\mathbf{B} = \mathrm{trace}(\mathbf{B}^\mathrm{T} \mathbf{A})$ \cite{zadeh2017multi}.

Combining this new definition of the \index{fourth-order diffusion tensor} fourth-order diffusion tensor with \index{Dirichlet boundary condition} Dirichlet boundary conditions as in Equation~(\ref{eq:inpaintingModel}) results in our proposed model:
\begin{equation} \label{eq:genModel}
\begin{split}
	{\partial_t u} = &- \partial_{xx}[\mathcal{D}(\mathbf{J}(u_\sigma)):\mathbf{H}(u)]_{xx} - \partial_{yx}[\mathcal{D}(\mathbf{J}(u_\sigma)):\mathbf{H}(u)]_{xy} \  \\
&	- \partial_{xy}[\mathcal{D}(\mathbf{J}(u_\sigma)):\mathbf{H}(u)]_{yx} - \partial_{yy}[\mathcal{D}(\mathbf{J}(u_\sigma)):\mathbf{H}(u)]_{yy}, \quad \Omega \backslash K \times (0,\infty)\ , \\
	u = &\, f, \quad K\times [0,\infty)
\end{split}
\end{equation}

As it is customary in PDE-based inpainting, we allow Equation~(\ref{eq:genModel}) to evolve until a steady state has been reached, i.e., the time derivative becomes negligible. In our numerical implementation, we use a \index{Fast Semi-Iterative Scheme} Fast Semi-Iterative Scheme (FSI) \cite{hafner2016fsi} to greatly  accelerate convergence to a large stopping time.

In the definition of our \index{fourth-order diffusion tensor} fourth-order diffusion tensor $\mathcal{D}$, the choice of $\mu_1$ and $\mu_2$ is analogous to anisotropic edge enhancing diffusion \cite{weickert1998anisotropic}. However, two additional terms occur in the fourth-order case, $\mu_3$ and $\mu_4$. As noted in \cite{zadeh2017multi}, $\mu_4$ is irrelevant, since the corresponding eigentensor $\mathbf{E}_4$ is anti-symmetric, and the dot product $\mathbf{E}_4:\mathbf{H}$ with the Hessian of any sufficiently smooth image will be zero due to its symmetry. To better understand the role of $\mu_3$, we observe that
\begin{equation} \label{eq:E3}
\begin{split}
  \mathbf{E}_3:\mathbf{H}&=\frac{1}{\sqrt{2}}\left(\mathbf{v}_1^\mathrm{T} \mathbf{H} \mathbf{v}_2 + \mathbf{v}_2^\mathrm{T} \mathbf{H} \mathbf{v}_1\right) \\
&=\frac{1}{\sqrt{2}} \left(
u_{\big(\frac{\mathbf{v}_1+\mathbf{v}_2}{\sqrt{2}}\big)\big(\frac{\mathbf{v}_1+\mathbf{v}_2}{\sqrt{2}}\big)}
-u_{\big(\frac{\mathbf{v}_1-\mathbf{v}_2}{\sqrt{2}}\big)\big(\frac{\mathbf{v}_1-\mathbf{v}_2}{\sqrt{2}}\big)}\right)\ ,
\end{split}
\end{equation}
which amounts to a mixed second derivative of $u$, in directions along and orthogonal to the regularized image gradient $\frac{\nabla u_{\sigma}}{||\nabla u_{\sigma}||_2}$ or, equivalently, to the difference of second derivatives in the directions that are exactly in between the two. This term vanishes if the Hessian is isotropic, or if the gradient is parallel to one of the Hessian eigenvectors. Therefore, the role of $\mu_3$ can be seen as steering the amount of diffusion in cases of a Hessian anisotropy that goes along with a misalignment between gradient and Hessian eigenvectors.

\begin{figure}[tbp]
  \centering
  \begin{minipage}{0.31\linewidth}
    \includegraphics[width=\linewidth]{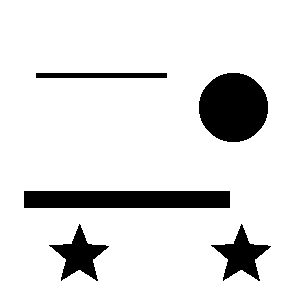}
    
    \footnotesize (a) Original test image of size $300\times 300$
  \end{minipage}
  \hskip0.02\linewidth
  \begin{minipage}{0.31\linewidth}
    \includegraphics[width=\linewidth]{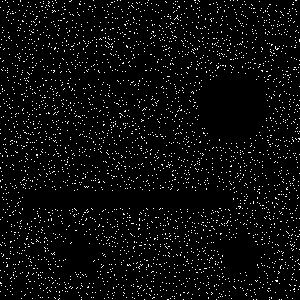}
    
    \footnotesize (b) Randomly chosen $5\%$ of pixel values
  \end{minipage}
  \hskip0.02\linewidth
  \begin{minipage}{0.31\linewidth}
    \includegraphics[width=\linewidth]{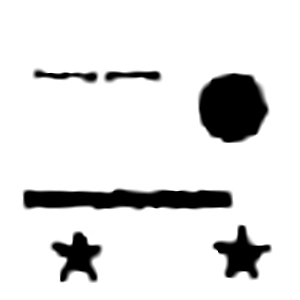}
    
    \footnotesize (c) Second-order EED inpainting based on (b)
  \end{minipage}

  \begin{minipage}{0.31\linewidth}
    \includegraphics[width=\linewidth]{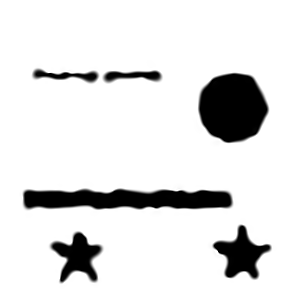}
    
    \footnotesize (d) Fourth-order EED inpainting with $\mu_3 = 1$
  \end{minipage}
  \hskip0.02\linewidth
  \begin{minipage}{0.31\linewidth}
    \includegraphics[width=\linewidth]{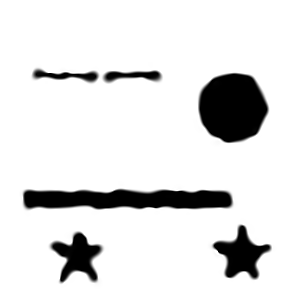}
    
    \footnotesize (e) Fourth-order EED inpainting with $\mu_3 = \frac{\mu_1+\mu_2}{2}$
  \end{minipage}
  \hskip0.02\linewidth
  \begin{minipage}{0.31\linewidth}
    \includegraphics[width=\linewidth]{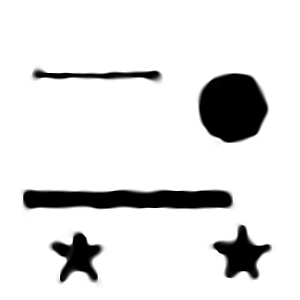}
    
    \footnotesize (f) Fourth-order EED inpainting with $\mu_3 = \sqrt{\mu_1 \mu_2}$
  \end{minipage}
  \caption{\label{fig:synth_test_image}%
    Reconstruction of a synthetic test image~(a) from 5\% of its pixels~(b) based on second-order diffusion~(c) and fourth-order diffusion with different coefficients for the mixed term $\mu_3$ (d--f). Visually, the reconstruction in (f) is most similar to the original image.}
\end{figure}

Gorgi Zadeh et al.\ \cite{zadeh2017multi} simply set $\mu_3$ to the arithmetic mean of $\mu_1$ and $\mu_2$. In our work, we empirically evaluated several alternative options for $\mu_3$ by reconstructing the test image shown in Figure~\ref{fig:synth_test_image}~(a), which contains one rectangle, one circle, and two stars, from a randomly selected subset of $5\%$ of its pixels. In this experiment, we compare EED based inpainting with our novel \index{fourth-order edge enhancing diffusion (FOEED)} fourth-order edge enhancing diffusion (FOEED) with different settings of $\mu_3$: Specifically, $\mu_3=1$ corresponds to the maximum of $\mu_1$ and $\mu_2$ (Figure~\ref{fig:synth_test_image}~(d)),  $\mu_3=\left(1+g(\lambda_1)\right)/2$ corresponds to their arithmetic mean (Figure~\ref{fig:synth_test_image}~(e)), and $\mu_3=\sqrt{g(\lambda_1)}$ corresponds to their geometric mean (Figure~\ref{fig:synth_test_image}~(f)). In all cases, we used the Charbonnier diffusivity (Equation~(\ref{eq:CharDiff})), which is popular for image compression \cite{galic2008image}, the same contrast parameter ($\lambda = 0.1$) and smoothing parameter ($\sigma = 1$). The only difference is time step size, where second-order EED permitted a stable step size of $0.25$, while a smaller step size of $0.05$ was chosen for FOEED. A more detailed theoretical and empirical analysis of stability will be given in Section~\ref{subsec:implementation}.

A numerical comparison of the results is given in Table~\ref{tab:syntezImg_errors}. For evaluation, we used the well-known \index{mean squared error (MSE)} mean squared error (MSE) and \index{average absolute error (AAE)} average absolute error (AAE) between original and reconstructed images. For two-dimensional gray-valued images $u$ and $v$ with the same dimensions $m\times n$, the MSE and AAE are defined as
\begin{equation}
\begin{split}
	\mathrm{MSE}(u,v) &= \frac{1}{mn}\sum_{i,j}(u_{i,j}-v_{i,j})^2 \ , \\
	\mathrm{AAE}(u,v) &= \frac{1}{mn}\sum_{i,j}|u_{i,j}-v_{i,j}| \ .
\end{split}
\end{equation}

According to Table~\ref{tab:syntezImg_errors}, the most accurate results are achieved by setting $\mu_3$ to the geometric mean of $\mu_1$ and $\mu_2$. Fourth-order EED with this setting produces higher accuracy than second-order EED. Visually, Figure~\ref{fig:synth_test_image} supports this conclusion. Specifically, fourth-order EED with $\mu_3=\sqrt{\mu_1 \mu_2}$ is the only variant that correctly connects the thin bar at the top of the test image, and it leads to a straighter shape of the thicker bar below, which is more similar to its original rectangular shape. In all subsequent experiments, we set $\mu_3=\sqrt{\mu_1 \mu_2}$.

\begin{table}[t]
\caption{Numerical reconstruction errors on the test image (Figure~\ref{fig:synth_test_image})}
\label{tab:syntezImg_errors}       % Give a unique label
\begin{tabular}{p{1.4cm}p{1.4cm}p{2.5cm}p{2.9cm}p{2.9cm}}
\hline\noalign{\smallskip}
Errors & EED & FOEED \footnotesize{($\mu_3=1$)} & FOEED \footnotesize{($\mu_3=\frac{\mu_1+\mu_2}{2}$)} & FOEED \footnotesize{($\mu_3=\sqrt{\mu_1 \mu_2}$)}\\
\noalign{\smallskip}\svhline\noalign{\smallskip}
MSE & 647.183 & 660.588 & 634.321 & 533.987\\
AAE  & 5.043 & 4.581 & 4.505 & 4.140\\
\noalign{\smallskip}\hline\noalign{\smallskip}
\end{tabular}
\end{table}

%------------------------------------------------------------------------
\subsection{A Unifying Framework for Fourth-Order Diffusion}
\label{subsec:comparison}
Several fourth-order diffusion PDEs have been used for image processing previously. We can better understand how they relate to our newly proposed PDE by observing that the fourth-order diffusion tensor $\mathcal{D}$ introduces a unifying framework for fourth-order diffusion filters. In particular, given its coefficients $d_{ijkl}$, we can expand Equation~(\ref{eq:ts-fourth-order-diffusion}) by using Einstein notation as
\begin{equation} \label{eq:gen4thPDE_Model}
	{\partial_t u} = - \partial_{ij}[d_{ijkl}u_{kl}] 
\end{equation}

Effectively, the fourth-order diffusion tensor allows us to separately set the diffusivities for all $2^4=16$ fourth-order derivatives of the two-dimensional image $u$. We will now demonstrate how several well-known fourth-order PDEs can be expressed in this framework, starting with the You-Kaveh PDE \cite{you2000fourth} 
\begin{equation} \label{eq:YK_Model}
	\partial_t u = - \Delta(g(|\Delta u|) \Delta u) \ ,
\end{equation}
which can be rewritten as
\begin{equation} \label{eq:YK4thPDE_Model}
\begin{split}
	{\partial_t u} = - \partial_{xx}[g(|\Delta u|)u_{xx} + 0\cdot u_{xy} + 0\cdot u_{yx} + g(|\Delta u|)u_{yy}] \\ 
		             - \partial_{yx}[0\cdot u_{xx} + 0\cdot u_{xy} + 0\cdot u_{yx} + 0\cdot u_{yy}] \\
	                 - \partial_{xy}[0\cdot u_{xx} + 0\cdot u_{xy} + 0\cdot u_{yx} + 0\cdot u_{yy}] \\
	                 - \partial_{yy}[g(|\Delta u|)u_{xx} + 0\cdot u_{xy} + 0\cdot u_{yx} + g(|\Delta u|)u_{yy}] \ .
\end{split}
\end{equation}

Here and in all subsequent examples, many terms have zero coefficients. For brevity, we will omit them from now on.

Hajiaboli's anisotropic fourth-order PDE \cite{hajiaboli2011anisotropic} is
\begin{equation} \label{eq:Haji_PDE}
	\partial_t u = - \Delta\left(g(||\nabla u||)^2u_{NN} + g(||\nabla u||)u_{TT}\right) \ ,
\end{equation}
where $N$ and $T$ are unit vectors parallel and orthogonal to the gradient, respectively. It can be rewritten as
\begin{equation} \label{eq:Haji4thPDE_Model}
\begin{split}
	{\partial_t u} = - \partial_{xx}\Bigg[\Bigg(\frac{g(||\nabla u||)^2u_x^2+g(||\nabla u||)u_y^2}{u_x^2+u_y^2}\Bigg)u_{xx} + \Bigg(\frac{g(||\nabla u||)^2u_xu_y-g(||\nabla u||)u_xu_y}{u_x^2+u_y^2}\Bigg) u_{xy} \\ 
	+ \Bigg(\frac{g(||\nabla u||)^2u_xu_y-g(||\nabla u||)u_xu_y}{u_x^2+u_y^2}\Bigg)  u_{yx} + \Bigg(\frac{g(||\nabla u||)^2u_y^2+g(||\nabla u||)u_x^2}{u_x^2+u_y^2}\Bigg)u_{yy}\Bigg] \\ 
%		             - \partial_{yx}[0\cdot u_{xx} + 0\cdot u_{xy} + 0\cdot u_{yx} + 0\cdot u_{yy}] \\
%	                 - \partial_{xy}[0\cdot u_{xx} + 0\cdot u_{xy} + 0\cdot u_{yx} + 0\cdot u_{yy}] \\
	                 - \partial_{yy}\Bigg[\Bigg(\frac{g(||\nabla u||)^2u_x^2+g(||\nabla u||)u_y^2}{u_x^2+u_y^2}\Bigg)u_{xx} + \Bigg(\frac{g(||\nabla u||)^2u_xu_y-g(||\nabla u||)u_xu_y}{u_x^2+u_y^2}\Bigg) u_{xy} \\ 
	+ \Bigg(\frac{g(||\nabla u||)^2u_xu_y-g(||\nabla u||)u_xu_y}{u_x^2+u_y^2}\Bigg)  u_{yx} + \Bigg(\frac{g(||\nabla u||)^2u_y^2+g(||\nabla u||)u_x^2}{u_x^2+u_y^2}\Bigg)u_{yy}\Bigg]
\end{split}
\end{equation}

From this method, Li et al.\ \cite{li2013two} derived two anisotropic fourth-order PDEs that, to our knowledge, are the only anisotropic fourth-order models that have been applied to inpainting previously. We will refer to them as Li~1
\begin{equation} \label{eq:li1_Model}
	\partial_t u = -\Delta(g(||\nabla u||)u_{NN}+u_{TT})
\end{equation}
and Li~2
\begin{equation} \label{eq:li2_Model}
	\partial_t u = -\Delta(u_{TT}) \ .
\end{equation}

Li~1 can be re-written as
\begin{equation} \label{eq:li1_GenModel}
\begin{split}
	\partial_t u = -\partial_{xx}\Bigg[\Bigg(\frac{g(||\nabla u||)u_x^2+u_y^2}{u_x^2+u_y^2}\Bigg)u_{xx} + \Bigg(\frac{g(||\nabla u||)u_xu_y-u_xu_y}{u_x^2+u_y^2}\Bigg) u_{xy} \\ 
	+ \Bigg(\frac{g(||\nabla u||)u_xu_y-u_xu_y}{u_x^2+u_y^2}\Bigg)u_{yx} + \Bigg(\frac{g(||\nabla u||)u_y^2+u_x^2}{u_x^2+u_y^2}\Bigg)u_{yy}\Bigg] \\
%	- \partial_{yx}[0\cdot u_{xx} + 0\cdot u_{xy} + 0\cdot u_{yx} + 0\cdot u_{yy}] \\
%    - \partial_{xy}[0\cdot u_{xx} + 0\cdot u_{xy} + 0\cdot u_{yx} + 0\cdot u_{yy}] \\
    -\partial_{yy}\Bigg[\Bigg(\frac{g(||\nabla u||)u_x^2+u_y^2}{u_x^2+u_y^2}\Bigg)u_{xx} + \Bigg(\frac{g(||\nabla u||)u_xu_y-u_xu_y}{u_x^2+u_y^2}\Bigg) u_{xy} \\ 
	+ \Bigg(\frac{g(||\nabla u||)u_xu_y-u_xu_y}{u_x^2+u_y^2}\Bigg)u_{yx} + \Bigg(\frac{g(||\nabla u||)u_y^2+u_x^2}{u_x^2+u_y^2}\Bigg)u_{yy}\Bigg]\, ,
\end{split}
\end{equation} while Li~2 becomes
\begin{equation} \label{eq:li2_GenModel}
\begin{split}
	\partial_t u = -\partial_{xx}\Bigg[\Bigg(\frac{u_y^2}{u_x^2+u_y^2}\Bigg)u_{xx} + \Bigg(\frac{-u_xu_y}{u_x^2+u_y^2}\Bigg) u_{xy} 
	+ \Bigg(\frac{-u_xu_y}{u_x^2+u_y^2}\Bigg)u_{yx} + \Bigg(\frac{u_x^2}{u_x^2+u_y^2}\Bigg)u_{yy}\Bigg] \\
%	- \partial_{yx}[0\cdot u_{xx} + 0\cdot u_{xy} + 0\cdot u_{yx} + 0\cdot u_{yy}] \\
%    - \partial_{xy}[0\cdot u_{xx} + 0\cdot u_{xy} + 0\cdot u_{yx} + 0\cdot u_{yy}] \\
    -\partial_{yy}\Bigg[\Bigg(\frac{u_y^2}{u_x^2+u_y^2}\Bigg)u_{xx} + \Bigg(\frac{-u_xu_y}{u_x^2+u_y^2}\Bigg) u_{xy} 
	+ \Bigg(\frac{-u_xu_y}{u_x^2+u_y^2}\Bigg)u_{yx} + \Bigg(\frac{u_x^2}{u_x^2+u_y^2}\Bigg)u_{yy}\Bigg]\, .
\end{split}
\end{equation}

We observe that Li~1 is based on a similar idea as our proposed PDE: It permits \index{fourth-order diffusion} fourth-order diffusion along the edge, while applying a nonlinear \index{diffusivity function} diffusivity function across the edge. However, expressing Li et al.'s models in terms of fourth-order diffusion tensors $\mathcal{D}_1$ and $\mathcal{D}_2$ reveals that our approach is more general. In particular, we can observe that
\begin{equation} \label{eq:li1li2_diff_direc}
\begin{split}
	\mathcal{D}_1:\mathbf{H} &= g(||\nabla u||)u_{NN} \mathbf{I} + u_{TT} \mathbf{I} \ ,\\
	\mathcal{D}_2:\mathbf{H} &= u_{TT} \mathbf{I} \ ,
\end{split}
\end{equation} 
where $\mathbf{I}$ is the $2\times 2$ identity matrix. In our model, $\mathcal{D}:\mathbf{H}$ can yield arbitrary anisotropic tensors. In this sense, our model more fully accounts for anisotropy compared to the ones by Hajiaboli and Li et al.

The fourth-order Equation~(\ref{eq:gen4thPDE_Model}) involves inner second derivatives of the image, which then get scaled by diffusivities, before outer second derivatives are taken. We observe that, in both cases, our model accounts for mixed derivatives that are ignored by previous approaches to anisotropic fourth-order diffusion: In the outer derivatives, this can be seen from the fact that Eq.~(\ref{eq:ts-fourth-order-diffusion}) involves mixed derivatives, while Hajiaboli and Li et al.\ only consider the Laplacian. 

Similarly, our definition of a \index{fourth-order diffusion tensor} fourth-order diffusion tensor $\mathcal{D}$ accounts for mixed derivatives also in the inner derivatives. Following Equation~(\ref{eq:E3}), we obtain
\begin{equation} \label{eq:foad_diff_direc}
\begin{split}
	\mathcal{D}:\mathbf{H} &= \mu_1(\mathbf{E}_1\otimes \mathbf{E}_1):\mathbf{H} + \mu_2(\mathbf{E}_2\otimes \mathbf{E}_2):\mathbf{H} + \mu_3(\mathbf{E}_3\otimes \mathbf{E}_3):\mathbf{H} \\
			     &= \mu_1 u_{\mathbf{v}_1\mathbf{v}_1} \mathbf{E}_1 + \mu_2 u_{\mathbf{v}_2\mathbf{v}_2} \mathbf{E}_2 + \frac{\mu_3}{\sqrt{2}}\left( u_{\big(\frac{\mathbf{v}_1+\mathbf{v}_2}{\sqrt{2}}\big)\big(\frac{\mathbf{v}_1+\mathbf{v}_2}{\sqrt{2}}\big)} - u_{\big(\frac{\mathbf{v}_1-\mathbf{v}_2}{\sqrt{2}}\big)\big(\frac{\mathbf{v}_1-\mathbf{v}_2}{\sqrt{2}}\big)}\right) \mathbf{E}_3 \ . \\
\end{split}
\end{equation} 

Comparing Equations (\ref{eq:li1li2_diff_direc}) and (\ref{eq:foad_diff_direc}) reveals differences in the considered directions: First, $N$ and $T$ are derived from the unregularized gradient, while the corresponding directions $\mathbf{v}_1$ and $\mathbf{v}_2$ in our model include a \index{Gaussian pre-smoothing} Gaussian pre-smoothing. A second difference is that our model involves an additional term, which is steered by $\mu_3$, and accounts for the directions in between the regularized gradient and its orthogonal vectors, i.e., $\big(\frac{\mathbf{v}_1+\mathbf{v}_2}{\sqrt{2}}\big)$ and $\big(\frac{\mathbf{v}_1-\mathbf{v}_2}{\sqrt{2}}\big)$. As it was demonstrated in the previous section, this term can have a noticeable effect on the outcome. Overall, we conclude that our newly proposed model is more general than the previously published ones.

\subsection{Discretization and Stability}
\label{subsec:implementation}
When discretizing Equation \eqref{eq:genModel} with standard \index{finite differences} finite differences
\begin{equation}
\begin{split}
	u_{xx} &\approx \frac{(u_{i-1,j}-2u_{i,j}+u_{i+1,j})}{(\Delta x)^2} \ , \\
	u_{yy} &\approx \frac{(u_{i,j-1}-2u_{i,j}+u_{i,j+1})}{(\Delta y)^2} \ , \\
	u_{xy} &\approx \frac{(u_{i-1,j-1}+u_{i+1,j+1}+u_{i-1,j+1}+u_{i+1,j-1})}{4(\Delta x)(\Delta y)} \ , \\
	u_{yx} &= u_{xy}\ ,
\end{split}
\end{equation}
we can write it down in matrix-vector form as in \cite{zadeh2017multi},
\begin{equation} \label{eq:discrtModel}
	\mathbf{u}^{k+1} = \mathbf{u}^{k}(\mathbf{I}-\tau \  \mathbf{P}_k) \ ,
\end{equation}
where $u^{k}$ is an $mn$ dimensional image vector at iteration $k$. $m, n$ are image width and height respectively; $\Delta x$ and $\Delta y$ are the corresponding pixel edge lengths. $\mathbf{P}_k$ is a positive semi-definite matrix that, with step size $\tau$, leads to the system matrix ($\mathbf{I}-\tau \ \mathbf{P}_k$). The notation $\mathbf{P}_k$ indicates that it is iteration dependent, i.e., $\mathbf{P}_k = \mathbf{P}(\mathbf{u}_k)$. 

Stability of fourth-order PDEs for image processing is typically formalized in an $L_2$ sense, i.e., a time step is chosen such that
\begin{equation} \label{eq:l2stability}
	||\mathbf{u}^{k+1}||_2 \leq ||\mathbf{u}^{k}||_2 \ .
\end{equation}

In an inpainting scenario, it depends on our initialization of the unknown pixels whether we can expect Equation~(\ref{eq:l2stability}) to hold. Therefore, we rely on a \index{stability analysis} stability analysis of the smoothing variant of our proposed PDE. This variant is obtained by removing the \index{Dirichlet boundary condition} Dirichlet boundary conditions and instead solving a standard initial value problem. In this case, the stability analysis presented by Gorgi Zadeh et al. \cite{zadeh2017multi} carries over. It ensures that time step sizes
\begin{equation} \label{eq:timeStepBound}
	\tau \leq \frac{2}{16(\Delta x)^2 + 16(\Delta y)^2 + 2(\Delta x \Delta y)}
\end{equation}
are stable in the $L_2$ sense. For a spatial \index{discretization} discretization $\Delta x = \Delta y = 1$, this yields $\tau \leq 1/17 \approx 0.0588$. In inpainting, we empirically obtained a useful \index{steady state} steady state with a time step size $\tau \leq 0.066$, independent of the initialization. The fact that this slightly exceeds the theoretical step size reflects the fact that Equation~(\ref{eq:timeStepBound}) results from deriving a sufficient, not a necessary condition for stability.

Stability of fourth-order schemes generally requires a quite small time step $\tau$. This makes it computationally expensive to reach the steady state by evaluating Equation~(\ref{eq:discrtModel}). Hafner et al.\ \cite{hafner2016fsi} propose a remedy to this problem, the so-called \index{Fast Semi-Iterative Scheme (FSI)} Fast Semi-Iterative Scheme (FSI). It extrapolates the basic solver iteration with the previous iterate and serves as an accelerated explicit scheme. The acceleration of the explicit scheme \eqref{eq:discrtModel} is given as
\begin{equation}
	\mathbf{u}^{m,k+1} = \alpha_k \cdot (\mathbf{I}-\tau \mathbf{P}(\mathbf{u}^{m,k}))\mathbf{u}^{m,k} + (1-\alpha_k)\cdot \mathbf{u}^{m,k-1} \ , 
\end{equation}
where $ \mathbf{u}^{m,-1} := \mathbf{u}^{m,0}$ and $\alpha_k = \frac{4k+2}{2k+3}$ for $k = 0, \cdots , n-1$.
Here $m$ stands for outer cycle, i.e. $m$-th cycle with inner cycle of length $n$. And for passing to the next outer cycle, we set $\mathbf{u}^{m+1,0} := \mathbf{u}^{m,n}$. The \index{stability analysis} stability analysis requires the matrix $\mathbf{P}$ to be symmetric. This is satisfied since the \index{diffusion tensor} diffusion tensor $\mathcal{D}$ is symmetric, and symmetric central discretizations are used. In our implementation, we used $n=40$, and stopped iterating after the first outer cycle for which $\|\mathbf{u}^{m}-\mathbf{u}^{m-1}\|_2<10^{-4}$.

\section{Experimental Results}

To establish the usefulness of our proposed new model, we applied it to the reconstruction of images from a sparse subset of pixels (Section~\ref{subsec:reconstruction}). Moreover, we evaluate performance for a more classic inpainting task, scratch removal (Section~\ref{subsec:inpainting}). We also demonstrate how results depend on the chosen \index{diffusivity function} diffusivity function and \index{contrast parameter} contrast parameter (Section~\ref{subsec:parameter-study}).

%-------------------------------------------------------------------------
\subsection{Reconstruction From a Sparse Set of Pixels}
\label{subsec:reconstruction}

Improving image reconstruction from a sparse set of known pixels was the main motivation behind our work. Therefore, we applied it to two well-known natural images, \emph{lena} and \emph{peppers,} as well as to a medical image, a slice of a $T_1$ weighted brain MR scan \emph{(t1slice).} For \emph{lena,} we kept a random subset of only $2\%$ of the pixels. Due to the lower resolution of the \emph{peppers} and \emph{t1slice} images, we kept $5\%$ and $20\%$, respectively.

In all three cases, results from our approach (\index{FOEED} FOEED) were compared to results from second-order \index{EED} EED, as well as from the two anisotropic fourth-order PDEs proposed by Li et al.\ \cite{li2013two}. In all experiments, we used the Charbonnier diffusivity function, we set the contrast parameter to $\lambda = 0.1$, and the \index{pre-smoothing parameter} pre-smoothing parameter to $\sigma = 1$.

\begin{figure*}[tbp]
  \centering
  \mbox{} \hfill
  \includegraphics[width=.49\linewidth]{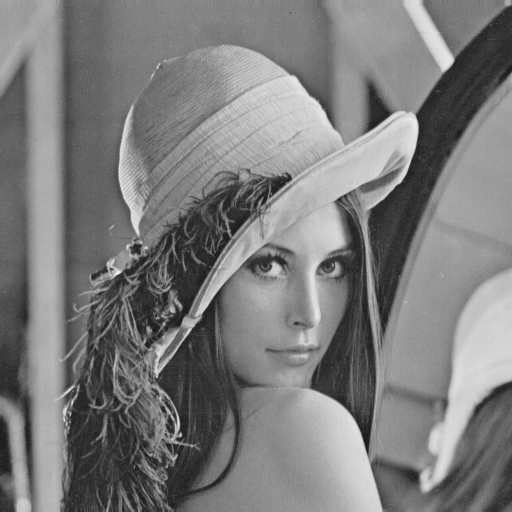}
  \hfill
  \includegraphics[width=.49\linewidth]{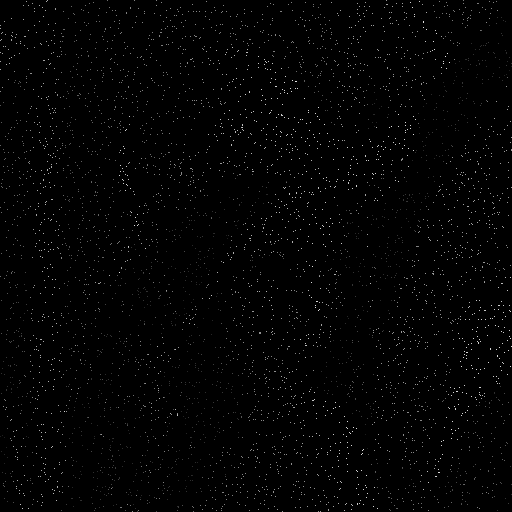}
  \hfill \mbox{}
  \centering
  \mbox{} \hfill
  \includegraphics[width=.49\linewidth]{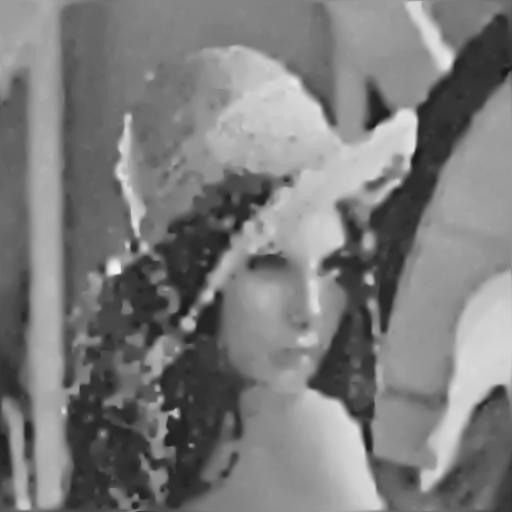}
  \hfill
  \includegraphics[width=.49\linewidth]{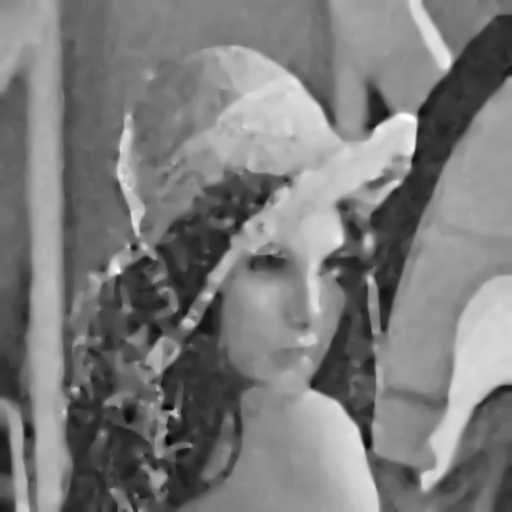}
  \hfill \mbox{}      
  \centering
  \mbox{} \hfill
  \includegraphics[width=.49\linewidth]{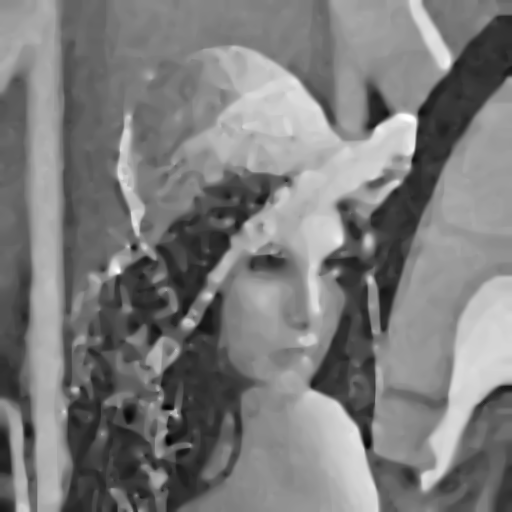}
  \hfill
  \includegraphics[width=.49\linewidth]{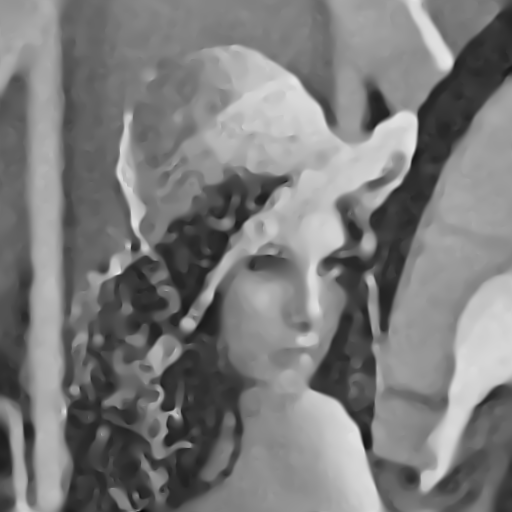}
  \hfill \mbox{}
  \caption{\label{fig:greyLena}%
	1st row left: original \textit{lena} image of size 512$\times$512; right: randomly chosen $2\%$ of pixel values;
	2nd row left: EED based inpainted image; right: Li1 based inpainted image;
	3rd row left: Li2 based inpainted image; right: FOEED based inpainted image;}	
\end{figure*}

\begin{figure*}[tbp]
  \centering
  \mbox{} \hfill
  \includegraphics[width=.49\linewidth]{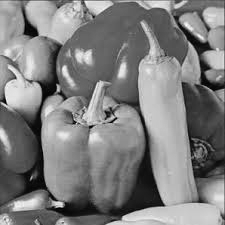}
  \hfill
  \includegraphics[width=.49\linewidth]{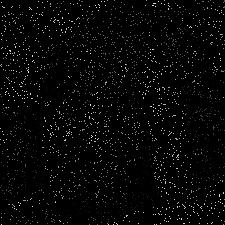}
  \hfill \mbox{}
  \centering
  \mbox{} \hfill
  \hfill
  \includegraphics[width=.49\linewidth]{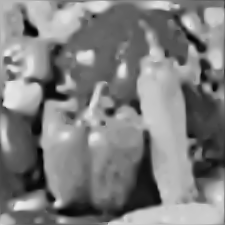}
  \hfill
  \includegraphics[width=.49\linewidth]{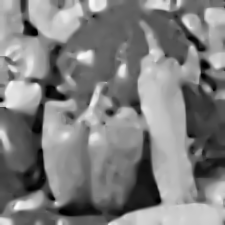}
  \hfill \mbox{}
  \centering
  \mbox{} \hfill
  \hfill
  \includegraphics[width=.49\linewidth]{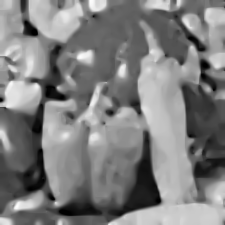}
  \hfill
  \includegraphics[width=.49\linewidth]{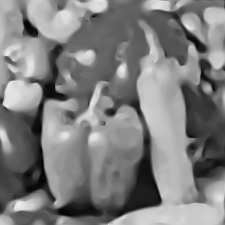}
  \hfill \mbox{}
  \caption{\label{fig:greyPepper}%
    1st row left: original \textit{peppers} image of size 225$\times$225; Right: randomly chosen $5\%$ of pixel values;
	2nd row left: EED based inpainted image; Right: Li1 based inpainted image;
	3rd row left: Li2 based inpainted image; Right: FOEED based inpainted image.}
\end{figure*}

\begin{figure*}[tbp]
  \centering
  \mbox{} \hfill
  \includegraphics[width=.49\linewidth]{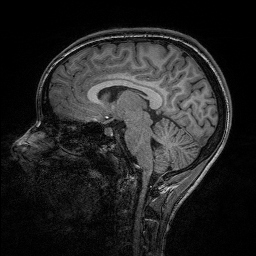}
  \hfill
  \includegraphics[width=.49\linewidth]{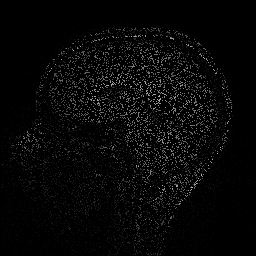}
  \hfill \mbox{}
  \centering
  \mbox{} \hfill
  \hfill
  \includegraphics[width=.49\linewidth]{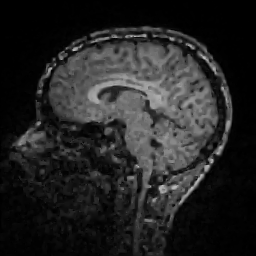}
  \hfill
  \includegraphics[width=.49\linewidth]{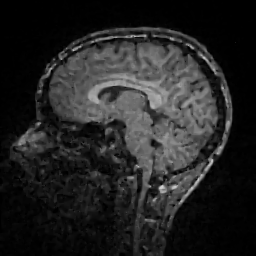}
  \hfill \mbox{}
  \centering
  \mbox{} \hfill
  \hfill
  \includegraphics[width=.49\linewidth]{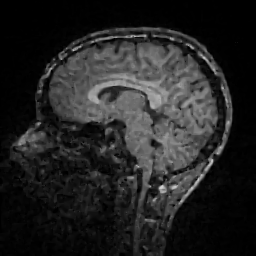}
  \hfill
  \includegraphics[width=.49\linewidth]{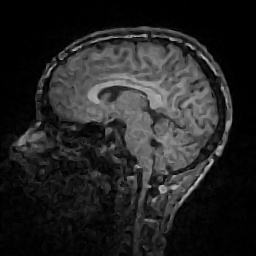}
  \hfill \mbox{}
  \caption{\label{fig:greyT1slice}%
    1st row left: original \textit{t1slice} image of size 256$\times$256; Right: randomly chosen $20\%$ of pixel values;
	2nd row left: EED based inpainted image; Right: Li1 based inpainted image;
	3rd row left: Li2 based inpainted image; Right: FOEED based inpainted image.}
\end{figure*}

Results for \emph{lena} are shown in Figure~\ref{fig:greyLena}, for \emph{peppers} in Figure~\ref{fig:greyPepper}, and for \emph{t1slice} in Figure~\ref{fig:greyT1slice}. A quantitative evaluation in terms of MSE and AAE is presented in Table~\ref{tab:greyImages}. In terms of the numerical results, our proposed method produced a more accurate reconstruction than any of the competing approaches. Visually, there is a clear difference between second-order (EED) and fourth-order approaches (Li1, Li2, FOEED). Especially, we found that the shapes of edges were reconstructed more accurately. For example, we noticed this around the shoulder and hat in the \textit{lena} image (Figure \ref{fig:greyLena}). Similarly, the white and grey matter boundaries were better separated in the \emph{t1slice} (Figure~\ref{fig:greyT1slice}).

As we expected based on the theoretical analysis in Section~\ref{subsec:comparison}, visual differences between the fourth-order methods are more subtle. However, in the \emph{peppers} image (Figure~\ref{fig:greyPepper}), the tall and thin and the small and thick peppers in the foreground are much more clearly separated in the FOEED result than in any of the others.

\begin{table}[t]
\caption{Numerical Comparison of Inpainting Models for Gray-Valued Images}
\label{tab:greyImages}       % Give a unique label
\begin{tabular}{p{1.5cm}p{1.5cm}p{2cm}p{2cm}p{2cm}p{2cm}}
\hline\noalign{\smallskip}
Image & Errors & EED & FOEED & Li1 & Li2\\
\noalign{\smallskip}\svhline\noalign{\smallskip}
\textit{lena} & MSE & 280.977 & \textbf{261.003} & 273.781 & 277.234\\
     & AAE & 8.511 & \textbf{8.226} & 8.551 & 8.638\\
\noalign{\smallskip}\hline\noalign{\smallskip}
\textit{peppers} & MSE & 467.261 & \textbf{443.129} & 455.633 & 459.606\\
     & AAE & 10.94 & \textbf{10.523} & 11.042 & 11.107\\     
\noalign{\smallskip}\hline\noalign{\smallskip}
\textit{t1-slice} & MSE & 166.356 & \textbf{150.002} & 152.698 & 155.955\\
     & AAE & 5.895 & \textbf{5.698} & 5.789 & 5.853\\     
\noalign{\smallskip}\hline\noalign{\smallskip}
\end{tabular}
\end{table}

\begin{figure*}
  \includegraphics[width=.49\linewidth]{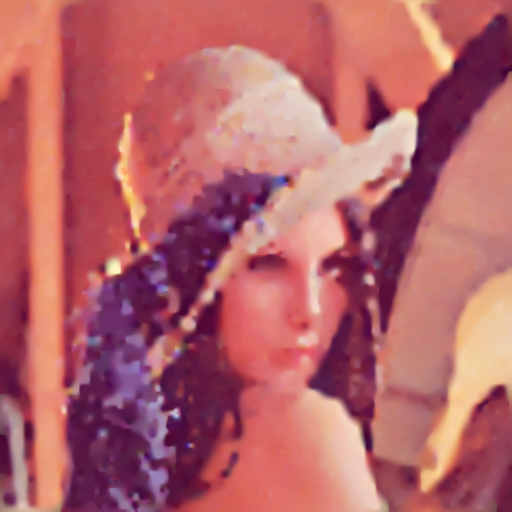}
  \hfill
  \includegraphics[width=.49\linewidth]{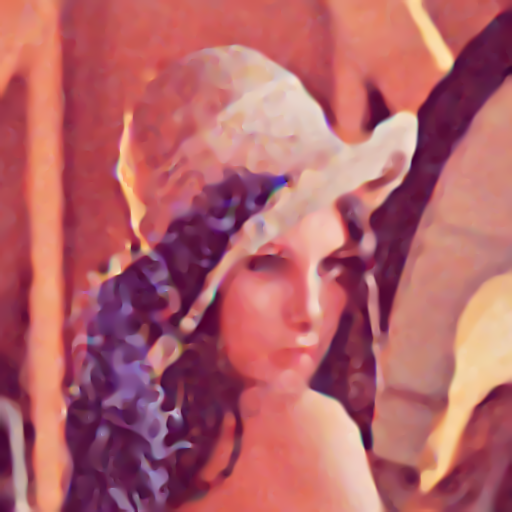}
  \caption{\label{fig:rgbLena}%
RGB \emph{lena} image, reconstructed from randomly chosen $2\%$ of pixel values using EED (left) or FOEED (right).}	
\end{figure*}

\begin{figure*}
  \includegraphics[width=.49\linewidth]{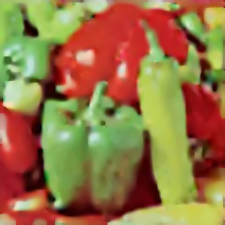}
  \hfill
  \includegraphics[width=.49\linewidth]{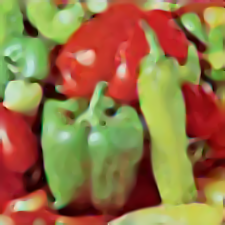}
  \caption{\label{fig:rgbPepper}%
RGB \emph{peppers} image, reconstructed from randomly chosen $5\%$ of pixel values using EED (left) or FOEED (right).} 
\end{figure*}

In addition to experimenting with grayscale versions of the \emph{lena} and \emph{peppers} images, we also applied \index{EED} EED and our \index{FOEED} FOEED filter channel-wise to the original RGB color versions. Results for \emph{lena} can be found in Figure~\ref{fig:rgbLena}, for \emph{peppers} in Figure~\ref{fig:rgbPepper}. Table~\ref{tab:rgbImages} again provides a quantitative comparison. Similar observations can be made as in the grayscale images: Again, FOEED leads to lower reconstruction errors than EED, it visually reconstructs edges more accurately, and separates the peppers more clearly.

\begin{table}[t]
\caption{Numerical Comparison of Inpainting Models for RGB Images}
\label{tab:rgbImages}       % Give a unique label
\begin{tabular}{p{2.5cm}p{2.5cm}p{3.1cm}p{3.1cm}}
\hline\noalign{\smallskip}
Image & Errors & EED & FOEED\\
\noalign{\smallskip}\svhline\noalign{\smallskip}
\textit{lena} & MSE & 283.585 & \textbf{260.003}\\
     & AAE & 8.529 & \textbf{8.207}\\
\noalign{\smallskip}\hline\noalign{\smallskip}
\textit{peppers} & MSE & 478.799 & \textbf{441.203}\\
     & AAE & 11.049 & \textbf{10.543}\\     
\noalign{\smallskip}\hline\noalign{\smallskip}
\end{tabular}
\end{table}

\begin{table}[t]
\caption{Numerical comparison and computation times corresponding to Figure~\ref{fig:greyImagesLargerMask}}
\label{tab:greyImagesLargerMask}       % Give a unique label
\begin{tabular}{p{2.0cm}p{2.3cm}p{2.3cm}p{2.3cm}p{2.3cm}}
\hline\noalign{\smallskip}
Image & Errors & EED & FOEED & CPU time\\
\noalign{\smallskip}\svhline\noalign{\smallskip}
\textit{lena} & MSE & 73.793 & \textbf{65.111} & 45.52 (FOEED)\\
     & AAE & 3.912 & \textbf{3.803} & 40.95 (EED)\\
\noalign{\smallskip}\hline\noalign{\smallskip}
\textit{peppers} & MSE & 113.5 & \textbf{110.885} & 20.999 (FOEED)\\
     & AAE & 4.565 & \textbf{4.441} & 19.79 (EED)\\     
\noalign{\smallskip}\hline\noalign{\smallskip}
\textit{t1-slice} & MSE & 114.845 & \textbf{107.323} & 24.74 (FOEED)\\
     & AAE & 4.610 & \textbf{4.553} & 10.64 (EED)\\     
\noalign{\smallskip}\hline\noalign{\smallskip}
\end{tabular}
\end{table}

\begin{figure*}[tbp]
  \centering
  \mbox{} \hfill
  \includegraphics[width=.49\linewidth]{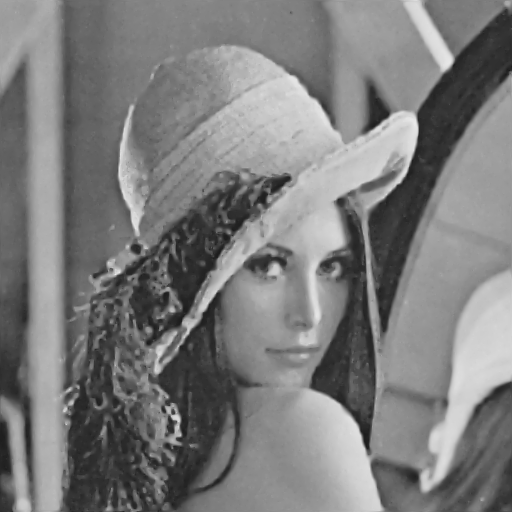}
  \hfill
  \includegraphics[width=.49\linewidth]{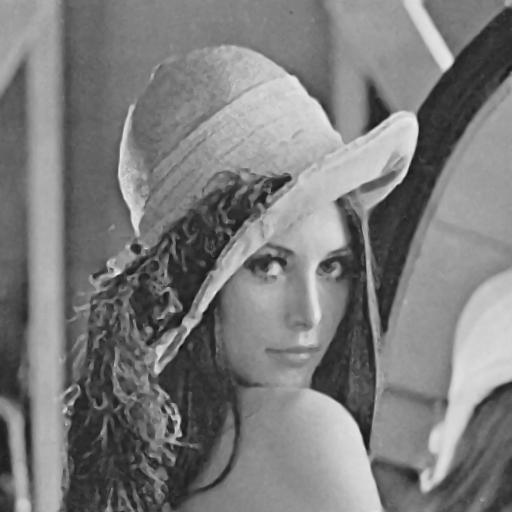}
  \hfill \mbox{}
  \centering
  \mbox{} \hfill
  \hfill
  \includegraphics[width=.49\linewidth]{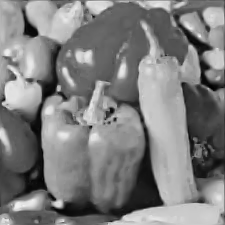}
  \hfill
  \includegraphics[width=.49\linewidth]{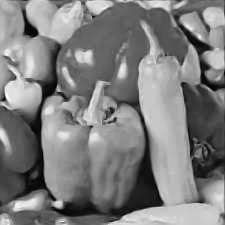}
  \hfill \mbox{}
  \centering
  \mbox{} \hfill
  \hfill
  \includegraphics[width=.49\linewidth]{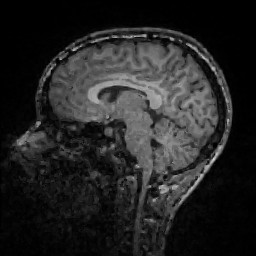}
  \hfill
  \includegraphics[width=.49\linewidth]{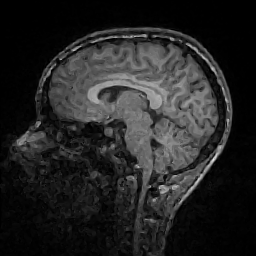}
  \hfill \mbox{}
  \caption{\label{fig:greyImagesLargerMask}%
    Higher quality reconstructions from a larger subset of pixels. 1st row: \textit{lena} image, reconstructed with EED (left) or FOEED (right) from randomly chosen $14\%$ of pixels;
	2nd row left: same for $20\%$ of pixels from \textit{peppers};
	3rd row left: same for $30\%$ of pixels from \textit{t1slice}. As expected, increasing the fraction of known pixels reduces the differences in the results of the two schemes.}
\end{figure*}

Finally, we reconstructed images from a larger number of pixels, to obtain visually cleaner results. Qualitative and numerical results are presented in Figure~\ref{fig:greyImagesLargerMask} and Table~\ref{tab:greyImagesLargerMask}, respectively. FOEED still yields lower numerical errors than EED. Unsurprisingly, the differences become smaller and less visually prominent as the mask density increases. The table also reveals that FOEED requires more CPU time compared to standard EED. However, due to the use of FSI in both cases, the difference in running times until convergence is much lower than the difference in time step sizes.

\subsection{Scratch Removal}
\label{subsec:inpainting}

\begin{figure*}[tbp]
  \centering
  \includegraphics[width=.49\linewidth]{pepper}
  \hfill
  \includegraphics[width=.49\linewidth]{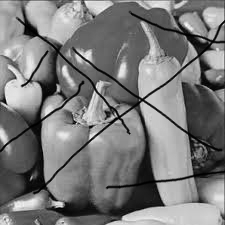}	  
  \centering
  \includegraphics[width=.49\linewidth]{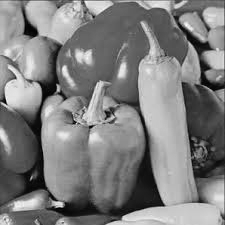}	  
  \hfill
  \includegraphics[width=.49\linewidth]{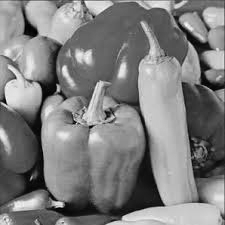}
  \centering
  \includegraphics[width=.49\linewidth]{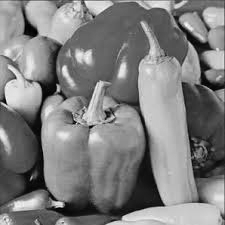}	  
  \hfill
  \includegraphics[width=.49\linewidth]{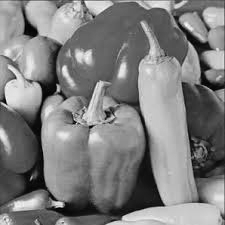}
  \caption{\label{fig:imgRestThin}%
	1st row left: original \textit{peppers} image of size 225$\times$225; Right: corrupted image.    
	2nd row left: EED based inpainting; Right: Li1 based inpainting. 
    3rd row left: Li2 based inpainting; Right: FOEED based inpainting.} 
\end{figure*}

Li et al.\ \cite{li2013two} proposed their anisotropic fourth-order PDE for more classical image inpainting tasks, such as scratch removal. We evaluated whether our more general filter can also provide a benefit in such a scenario by reconstructing a scratched version of the \emph{peppers} image. Similar to Li et al., we first made the scratches rather thin, covering only $6\%$ of all pixels. Results are shown in Figure~\ref{fig:imgRestThin} and in Table~\ref{tab:peppers_corr_thin}. In this case, all methods work well: Numerical errors are small and similar between methods, and even though FOEED achieves the best numerical result, differences are difficult to discern visually.

\begin{table}[tb]
\caption{Numerical comparison for peppers with thinner scratches (Figure~\ref{fig:imgRestThin})}
\label{tab:peppers_corr_thin}       % Give a unique label
\begin{tabular}{p{1.5cm}p{1.5cm}p{2cm}p{2cm}p{2cm}p{2cm}}
\hline\noalign{\smallskip}
Image & Errors & EED & FOEED & Li1 & Li2\\
\noalign{\smallskip}\svhline\noalign{\smallskip}
\textit{peppers} & MSE & 9.520 & \textbf{7.813} & 8.161 & 8.132\\
     & AAE & 0.363 & \textbf{0.326} & 0.346 & 0.346\\
\noalign{\smallskip}\hline\noalign{\smallskip}
\end{tabular}
\end{table}

\begin{figure*}[tbp]
  \centering
  \includegraphics[width=.49\linewidth]{pepper}
  \hfill
  \includegraphics[width=.49\linewidth]{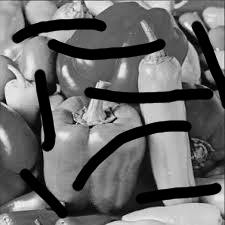}	  
  \centering
  \includegraphics[width=.49\linewidth]{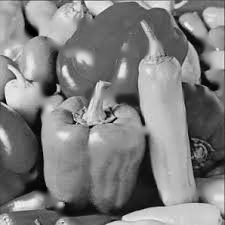}	  
  \hfill
  \includegraphics[width=.49\linewidth]{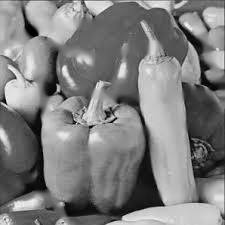}
  \centering
  \includegraphics[width=.49\linewidth]{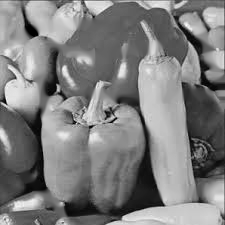}	  
  \hfill
  \includegraphics[width=.49\linewidth]{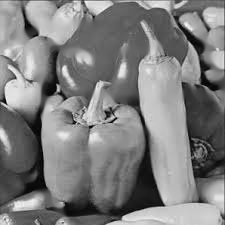}
  \caption{\label{fig:imgRest}%
	1st row left: original \textit{peppers} image of size 225$\times$225; Right: corrupted image.    
	2nd row left: EED based inpainting; Right: Li1 based inpainting. 
    3rd row left: Li2 based inpainting; Right: FOEED based inpainting.} 
\end{figure*}

Therefore, we created a more challenging version with thicker scratches, covering $18\%$ of all pixels (Figure~\ref{fig:imgRest}). The corresponding numerical comparison is shown in Table~\ref{tab:peppers_corr}. Here, FOEED achieves the most accurate reconstruction. Visually, we again observe that edges are reconstructed more accurately, and objects are more clearly separated, with fourth-order compared to second-order diffusion, and that steering it with a fourth-order diffusion tensor again provides small additional benefits over the previous methods.

\begin{table}[tb]
\caption{Numerical comparison for peppers with thicker scratches (Figure~\ref{fig:imgRest})}
\label{tab:peppers_corr}       % Give a unique label
\begin{tabular}{p{1.5cm}p{1.5cm}p{2cm}p{2cm}p{2cm}p{2cm}}
\hline\noalign{\smallskip}
Image & Errors & EED & FOEED & Li1 & Li2\\
\noalign{\smallskip}\svhline\noalign{\smallskip}
\textit{peppers} & MSE & 104.744 & \textbf{78.761} & 101.670 & 101.592\\
     & AAE & 2.455 & \textbf{2.099} & 2.465 & 2.450\\
\noalign{\smallskip}\hline\noalign{\smallskip}
\end{tabular}
\end{table}

\subsection{Effect of Diffusivity Function and Contrast Parameter}
\label{subsec:parameter-study}

For \index{image inpainting} image inpainting with \index{second-order PDEs} second-order PDEs, the Charbonnier diffusivity was previously found to work better than other established diffusivity functions. To assess whether this is still true in the fourth-order case, we repeated the reconstruction of the \emph{peppers} image as shown in Figure~\ref{fig:greyPepper} with different diffusivities. Table~\ref{tab:peppers_different_diffFuncs} summarizes the results. We conclude that the Charbonnier diffusivity still appears to be optimal.

\begin{figure*}[tbp]
  \centering
  \includegraphics[width=.32\linewidth]{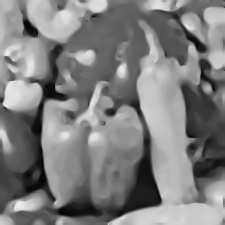}
\hfill 
  \includegraphics[width=.32\linewidth]{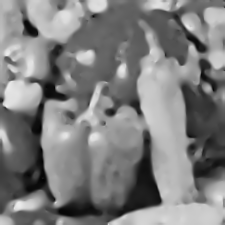}
  \hfill
  \includegraphics[width=.32\linewidth]{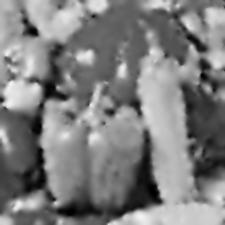}
  \caption{\label{fig:qGreyPepper_diff_lamdas}%
    From left to right: FOEED based inpainted image with $\lambda = 0.1$, $\lambda = 0.5$, $\lambda = 15.5$.}
\end{figure*}

Finally, in Figure \ref{fig:qGreyPepper_diff_lamdas}, we illustrate how the reconstructed image depends on the \index{contrast parameter} contrast parameter $\lambda$. As expected, increasing $\lambda$ leads to an increased blurring of edges. In the limit, the diffusivity function takes on values close to $1$ over a substantial part of the image, and our model starts to approximate homogeneous fourth-order diffusion.

\begin{table}[tb]
\caption{Numerical Comparison of FOEED with Different Diffusivity Functions}
\label{tab:peppers_different_diffFuncs}       % Give a unique label
\begin{tabular}{p{1.3cm}p{1.3cm}p{2.2cm}p{1.5cm}p{1.5cm}p{1.5cm}p{1.7cm}}
\hline\noalign{\smallskip}
Image & Errors & Charbonnier\cite{charbonnier1997deterministic} $\frac{1}{\sqrt{1+(\frac{s}{\lambda})^2}}$ & Aubert\cite{charbonnier1994two} $\frac{(\frac{s}{\lambda})^2}{(s^2+\lambda^2)^2}$ & Perona-Malik\cite{perona1990scale} $\frac{1}{1+(\frac{s}{\lambda})^2}$ & Perona-Malik2\cite{perona1990scale} $e^{-(\frac{s}{\lambda})^2}$ & Geman-Reynolds\cite{geman1992constrained} $\frac{2\lambda^2}{(s^2+\lambda^2)^2}$\\
\noalign{\smallskip}\svhline\noalign{\smallskip}
\textit{peppers} & MSE & \textbf{443.129} & 458.961 & 478.411 & 491.153 & 491.186\\
     & AAE & \textbf{10.523} & 10.587 & 10.943 & 11.157 & 11.007\\
\noalign{\smallskip}\hline\noalign{\smallskip}
\end{tabular}
\end{table}

\section{Conclusions}
We introduced a novel \index{fourth-order PDE} fourth-order PDE for \index{edge enhancing diffusion} edge enhancing diffusion (FOEED), steered by a \index{fourth-order diffusion tensor} fourth-order diffusion tensor. We implemented it using a fast semi-iterative scheme, and demonstrated that it achieved improved accuracy in several inpainting tasks, including reconstructing images from a small fraction of pixels, or removing scratches.

Our main motivation for using \index{fourth-order diffusion} fourth-order diffusion in this context is the increased smoothness of results compared to \index{second-order PDEs} second-order PDEs \cite{you2000fourth}, which we expected to result in visually more pleasant reconstructions. The model in our current work is still based on a single edge direction at each pixel, extracted via a traditional second-order \index{structure tensor} structure tensor. It is left as a separate research goal for future work to combine this with approaches for the estimation of complex structures such as crossings or bifurcations \cite{Aach:2006,Schultz:DagBook2009}, and with their improved reconstruction, e.g., by operating on the space of positions and orientations \cite{citti:2006,franken:2009,boscain:2014}.

Finally, our current work only considered reconstructions from a random subset of pixels. A practical image compression codec that uses our novel PDE should investigate how it interacts with more sophisticated approaches for selecting and coding inpainting masks \cite{schmaltz2014understanding}.
%-------------------------------------------------------------------------
\section*{Acknowledgement}
This research was supported by the German Academic Exchange Service (DAAD).
%-------------------------------------------------------------------------

\bibliographystyle{spmpsci}
\bibliography{chapter-5}
\end{document}